\documentclass[letterpaper, 10 pt, conference]{ieeeconf}  

\IEEEoverridecommandlockouts                              

\usepackage{graphics} 
\usepackage{epsfig} 
\usepackage{mathptmx} 
\usepackage{times} 
\usepackage{amsmath} 
\usepackage{amssymb}  
\usepackage{booktabs}
\usepackage{multirow}
\usepackage{algpseudocode}
\usepackage{float}
\usepackage{subfigure}
\usepackage[linesnumbered,ruled]{algorithm2e}
\usepackage{graphicx}
\usepackage{textcomp}
\usepackage{xcolor}
\usepackage{stackengine}

\usepackage{booktabs}
\usepackage{multirow}
\usepackage{algpseudocode}
\usepackage{float}
\usepackage{subfigure}
\usepackage[linesnumbered,ruled]{algorithm2e}
\usepackage{graphicx}
\usepackage{textcomp}
\usepackage{xcolor}
\usepackage{stackengine}
\usepackage{pgfplots}
\usepackage{pgfplotstable}
\usepackage{ifthen}
\pgfplotsset{compat=1.7}
\usepgfplotslibrary{groupplots}
\usepackage{verbatim}

\title{\LARGE \bf
Leveraging Explainability for Comprehending Referring Expressions\\in the Real World
}

\author{Fethiye Irmak Do\u{g}an$^{1}$, Gaspar I. Melsi\'{o}n$^{1}$ and Iolanda Leite$^{1}$
\thanks{$^{1}$Fethiye Irmak Do\u{g}an, Gaspar I. Melsi\'{o}n and Iolanda Leite are with the Division of Robotics, Perception and Learning from the School of Electrical Engineering and Computer Science at KTH Royal Institute of Technology, Stockholm, Sweden
        {\tt\small \{fidogan, gimp, iolanda\}@kth.se}}%
}

\begin{document}

\maketitle
\thispagestyle{empty}
\pagestyle{empty}

\begin{abstract}

For effective human-robot collaboration, it is crucial for robots to understand requests from users and ask reasonable follow-up questions when there are ambiguities. While comprehending the users' object descriptions in the requests, existing studies have focused on this challenge for limited object categories that can be detected or localized with existing object detection and localization modules. On the other hand, in the wild, it is impossible to limit the object categories that can be encountered during the interaction. To understand described objects and resolve ambiguities in the wild, for the first time, we suggest a method by leveraging explainability. Our method focuses on the active regions of a scene to find the described objects without putting the previous constraints on object categories and natural language instructions. We evaluate our method in varied real-world images and observe that the regions suggested by our method can help resolve ambiguities. When we compare our method with a state-of-the-art baseline, we show that our method performs better in scenes with ambiguous objects which cannot be recognized by existing object detectors.

\end{abstract}

\section{Introduction}



When humans and robots work on tasks as teammates, it is critical for robots to understand their human partners' natural language requests to successfully complete the task. During the task, the robot can encounter many challenges. For instance, when the robot is asked by its human partner to pick up an object, there can be misunderstandings caused by failures of speech recognition or the use of object descriptions that are unknown to the robot. 
Another challenge is an  ambiguous request (e.g., the human partner's object description might fit more than one object). 
In these cases, the robot should be able to ask efficient follow-up questions by using the familiar concepts in the request and making reasonable suggestions to its partner while asking for clarification. For example, it should suggest the objects that fit the description instead of just saying it couldn't understand the request.

People can identify objects with the help of referring expressions, which are phrases that describe the objects with their distinguishing features.
In robotics, comprehending object descriptions has been studied extensively. Prior work has focused on situated dialogue systems~\cite{zender2009situated, kruijff2007incremental}, probabilistic graph models~\cite{paul2016efficient}, and learning semantic maps~\cite{kollar2013learning}. Recent work on comprehending referring expressions has also employed models based on deep learning~\cite{hatori2018interactively,magassouba2019understanding,shridhar2018interactive,shridhar2020ingress}.

In this paper, we propose a method to comprehend users' expressions using deep neural networks' explainability in real-world, ambiguous environments. Although recent HRI studies evaluate the importance of explainable AI for different tasks~\cite{sridharan2019towards,edmonds2019tale,tabrez2019improving,siau2018building}, to our knowledge, this is the first work using explainability to comprehend the user descriptions. Recent models on comprehension of user expressions demonstrate promising results, but they assume the target candidates in a scene are given~\cite{magassouba2019understanding}, or these candidates can be obtained from the existing object detection~\cite{hatori2018interactively} or localization methods~\cite{shridhar2018interactive,shridhar2020ingress}. However, when robots are deployed in the real world, the encountered objects are not limited to the ones that can be detected by the state-of-art object detection or localization models, and it is not feasible to expand these models to localize every object category in a supervised fashion. Even when dealing with detectable object categories,  
due to environment conditions such as poor illumination or cluttered scenes these might not be possible to classify. 
In that case, when the described objects cannot be detected or localized, the existing solutions do not even consider these objects as target candidates. On the other hand, for a more general solution, our approach finds active regions of a scene using the explainability activations of an image captioning module, which is not trained on object-wise supervised fashion and learns a higher-level feature space 
-- see Section \ref{heatmap} for further information. Therefore,
our method does not require any detectable target candidates to suggest the described regions. This allows our system to handle various objects (including uncommon ones that may not be proposed by existing object detection or localization models) without putting any constraints on object categories or users' expressions. 

\begin{figure*}[t!]
\centering
\includegraphics[width=0.95\textwidth]{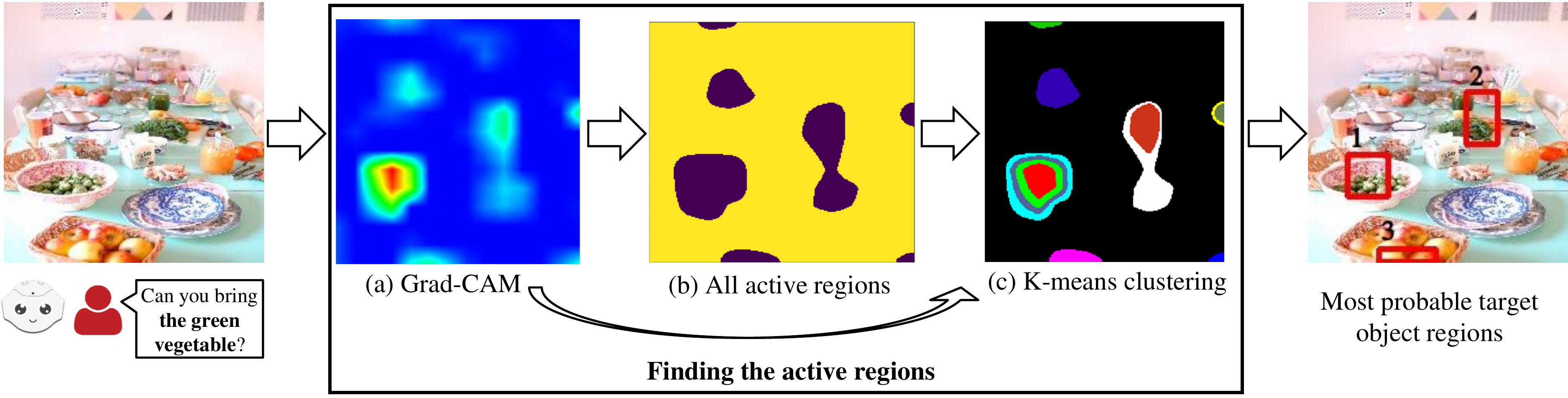}
\caption{Overview of our method to find the described object regions for a given scene and a referring expression (the bold part of the user's expression corresponds to the referring expression).}
\label{fig:overview}
\end{figure*}

In this work, we first find the active regions of a scene using the explainability module (i.e., Grad-CAM~\cite{selvaraju2017grad})
, and then we employ an unsupervised clustering technique (i.e., K-means) to find the active clusters. These active clusters are proposed as the regions that the robot needs to direct its attention to -- see Figure \ref{fig:overview} for the overview. When we examine the regions suggested by our method in varied real-world images, we observe that these regions can be useful for resolving ambiguities. Moreover, compared to a state-of-the-art referring expression comprehension model (i.e., MAttNet~\cite{yu2018MAttNet}), our method performs better in the scenes where several objects match the same description (e.g., multiple similar fishes) and where there are uncommon objects typically not recognized by off-the-shelf object detectors (e.g., an artichoke).


\subsection{Background}

Comprehension of referring expressions is of great importance in human-robot collaborative settings. %
Recent advances in this field have used multimodal strategies to combine different features from visual and language cues, including spatial relations~\cite{yu2018MAttNet,magassouba2019understanding}. %
It has been shown that methods with the ability to clarify ambiguous expressions through interaction improve the success rate \cite{hatori2018interactively, shridhar2020ingress}. %
Our proposed method is the first to employ explainability to solve the task of comprehending referring expressions while removing the dependency on using an object detection module that limits the results to the learned object categories. %
In this sense, the work from Shridhar et al.~\cite{shridhar2018interactive,shridhar2020ingress} avoided the use of predefined object categories as well, but it was still restricted to the target candidates obtained from the DenseCap object localization module~\cite{densecap}. Specifically, the work of Shridhar et al.\ proposed a two-stage approach where the first stage found visual descriptions of the candidate objects and the second stage obtained all pairwise relations between the candidates. Then, the obtained candidate object descriptions were compared with the user description to determine the most likely target candidate. In our work, we instead exploit the attention map obtained from an explainability module, focusing on regions of the image associated with the referring expression without constraining it to candidate objects.

Explainability has the objective of building models to be more transparent and understandable in their prediction-making process~\cite{BarredoArrieta2020}. It has been recognized to be crucial in the deployment of AI that impacts people's lives~\cite{Abdul2018,Bussone2015} as well as in the development of accountability in algorithms~\cite{Wieringa2020}. %
Kulesza et al.~\cite{Kulesza2013} showed that more complete explanations increased trust in automated decisions, although there is later evidence that this is only true with high-quality models~\cite{Smith-Renner2020}. %
Existing research in HRI has evaluated the importance of endowing robots with the ability to describe the reasoning behind their decisions to increase people's trust~\cite{siau2018building,edmonds2019tale} and improve human-robot collaboration~\cite{sridharan2019towards,setchi2020explainable}. %
Tabrez et al.~\cite{Tabrez2019explanation} demonstrated that a robot is perceived as more intelligent and helpful when it justifies its actions, and Edmonds et al.~\cite{edmonds2019tale} showed that symbolic explanation of the internal robot's decision process was more effective than a textual summary of it. %
As a novel aspect relative to previous work, we use an interpretability technique in the reverse order, where the explanations are used by the robot to find the correct region of an object described by a human.%

Recent computer vision studies have demonstrated the potential of interpretability to expand the use of explainability beyond the original concept of transparency by using explanations to improve models' intrinsic functioning. %
Hendricks et al.~\cite{Hendricks2018} used the saliency maps from Grad-CAM~\cite{selvaraju2017grad} (a method focused on providing CNNs with explainability for a given inference task) to force a captioning model to generate gender-specific words based on the person region of the image instead of the biased reasons given by gender-stereotyped datasets. %
Similarly, Ross et al.~\cite{Ross2017} improved model generalization by constraining explanations with input gradient penalties. %
Human attention maps have been aligned with the explanations provided by Grad-CAM to improve visual grounding in vision and language tasks~\cite{Selvaraju2019a}. %
Further, Li et al.~\cite{Li2018} presented a method to generate more accurate explanations (i.e., attention maps) through supervision in an end-to-end fashion while training the network.
Although our work does not use explainability during training, in line with enhancing the intrinsic functioning, our work leverages explainability 
to improve human-robot collaboration using Grad-CAM~\cite{selvaraju2017grad} saliency maps to direct the robot's attention to the appropriate regions described by the user -- see Section \ref{heatmap} for details.%

\subsection{Contributions}
The contributions of the paper are:
\begin{itemize}
    \item We propose using the explainability of image captioning to improve the effectiveness of referring expression comprehension. To our knowledge, this is the first work employing explainability for comprehending user expressions to direct robots to described objects in the wild, without any restrictions such as detectable or localizable objects.
    \item We examine the regions suggested by our method to determine whether these regions can be used for asking for clarification to resolve ambiguities.
    \item We compare our method with a state-of-the-art baseline in varied real-world images and show that our method performs better in challenging environments (i.e., scenes with uncommon and similar objects), which robots will more likely encounter in real-world.
\end{itemize}

\section{Finding the Described Scene Regions}
\label{Our}
For a given scene and a referring expression provided by a human using natural language, we aim to find the bounding boxes that show the described objects. To achieve this, we first use Grad-CAM~\cite{selvaraju2017grad} to find the active regions in the scene, and then we employ unsupervised clustering to find different clusters in these active regions. From these active clusters, we generate the bounding boxes most likely to belong to the target object regions (Figure \ref{fig:overview}).

\subsection{Finding the Active Regions}
\label{heatmap}

\begin{figure}
\centerline{
\subfigure[The input image]{
	\includegraphics[width=0.13\textwidth]{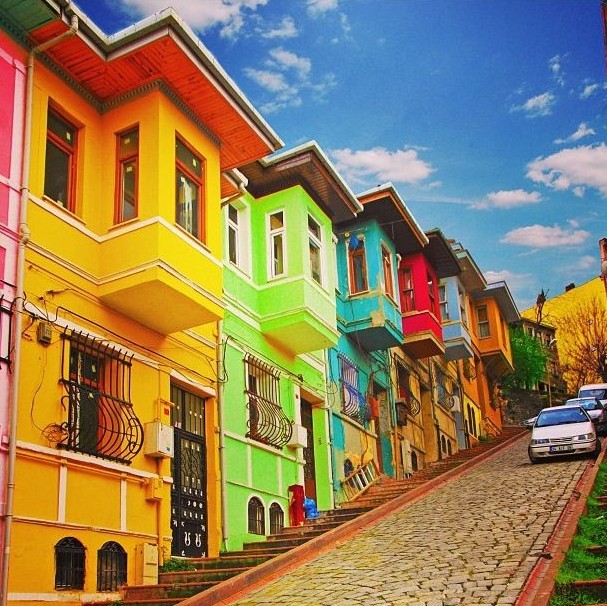}
	}
\subfigure[The heatmap]{
	\includegraphics[width=0.13\textwidth]{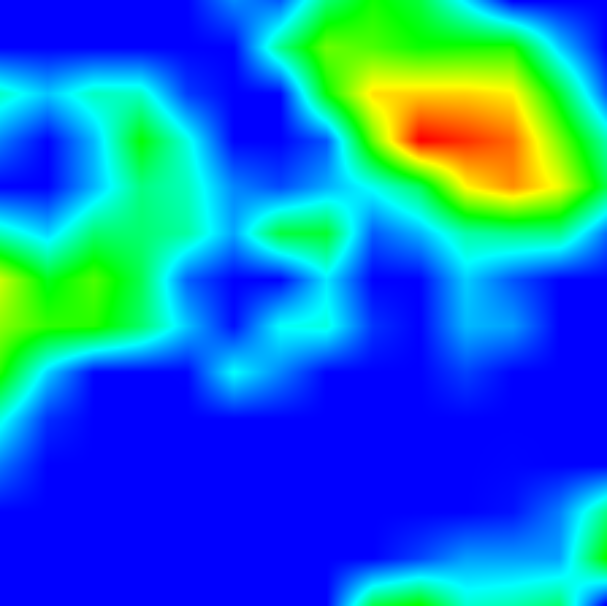}
	}
\subfigure[Active areas]{
	\includegraphics[width=0.13\textwidth]{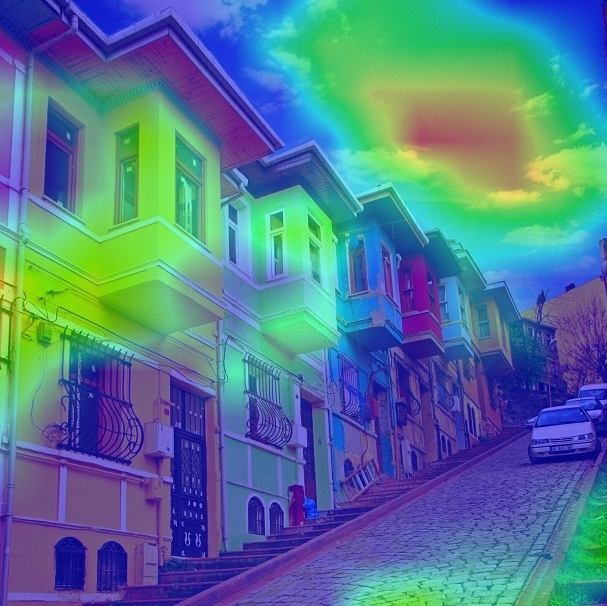}
	}
}
\caption{The heatmap from Grad-CAM (in (b)) and the activations aligned with the original image (in (c)) when the expression is `the blue sky'. The Grad-CAM heatmap highlights the sky using the color features, although the sky is not in the object categories of the MSCOCO dataset.}
\label{fig:mot}
\end{figure}

We use the image captioning module of Grad-CAM~\cite{selvaraju2017grad} to find active regions of a scene.
The module takes a scene and an expression as an input, and it generates a heatmap as an output. This heatmap shows the relevant regions in the scene. In order to obtain the heatmap, the module uses the NeuralTalk2 image captioning model~\cite{karpathy2015deep} and finds the gradient of the caption's log probability with respect to the final convolutional layer. Then, the module uses these gradients to provide visual explanations.

When different captions are provided for the same image, different regions become active depending on the items in the captions (e.g., different objects). In our work, these captions correspond to referring expressions, and we find the active regions specified by the referring expression (Figure \ref{fig:overview}(a)).

Using the NeuralTalk2 image captioning model with Grad-CAM has the advantage of not being restricted to specific object categories.  We achieve this because the NeuralTalk2 method was trained on a dataset (i.e., MSCOCO~\cite{lin2014microsoft} with five captions per image collected from crowd workers) that describes scenes with many different features, not restricted to object categories. Thanks to varied scene descriptions encountered during the training of NeuralTalk2, when an object category is unknown (i.e., not in MSCOCO object categories), the higher-level feature space learned by NeuralTalk2 and visualized by Grad-CAM can be used to show the active regions that fit the given description. For instance, in Figure \ref{fig:mot}, when the expression is `the blue sky', the highlighted region of Grad-CAM shows the sky, although the sky is not in the object categories of the MSCOCO. In that case, the color information is helpful for NeuralTalk2 to determine what to search for in the image. In this example, the existing works that first detect the candidate objects and select the target object among these candidates fail if they do not detect the sky, which is typically not recognized by off-the-shelf object detectors. On the other hand, by using the Grad-CAM activations of the NeuralTalk2 captioning method, we can consider the sky as a candidate region using the additional features given in the object description.

\subsection{Clustering Heatmaps}

After finding the active regions in a scene, we aim to cluster them. These clusters can be interpreted as different regions belonging to candidate objects so the robot can direct its attention to the right part of the scene. To achieve this, we first find the total number of active regions in the heatmap and use this value to determine the number of the resulting active clusters. Consequently, we employ K-means clustering to identify those clusters.

\begin{figure}
\centerline{
\subfigure[The input image]{
	\includegraphics[width=0.13\textwidth]{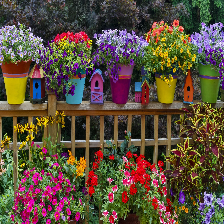}
	}
\subfigure[The heatmap]{
	\includegraphics[width=0.13\textwidth]{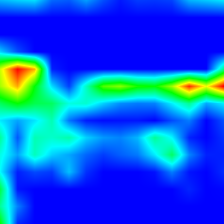}
	}
\subfigure[Active clusters]{
	\includegraphics[width=0.13\textwidth]{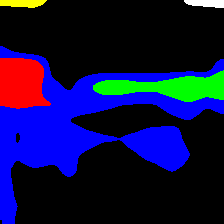}
	}
}
\caption{The heatmap from Grad-CAM (in (b)) and the active clusters from K-means (in (c)) when the expression is `the red house between the pink and yellow flower pots'. The activation heatmap is not straightforward to cluster without using K-means clustering.}
\label{fig:Clustering}
\end{figure}

\subsubsection{Finding the Number of Clusters}
\label{numCLus}
In order to determine the number of clusters in K-means clustering, we find the number of unconnected areas that are active in the heatmap~$\mathcal{H}$. We first define a set $\mathcal{U}$ where its values are 1 for active pixels and 0 otherwise:
%
\begin{equation}
\mathcal{U} = \{| r_p > T_h\ \textrm{or}\ g_p > T_h|, \quad \forall p \in \mathcal{H}\},
\label{eqn:num}
\end{equation}
{\noindent}where $|.|$ sets the value as 1 when the condition is correct, and 0 otherwise. Additionally, $r_p$ and $g_p$ show the normalized intensity values of each pixel $p$ for the red and green channels. We set the threshold $T_h$ as $0.9$ to only consider the pixels with high activation. A smaller value of this threshold can drastically increase the number of clusters by considering low-activation regions. With our formulation, $\mathcal{U}$ corresponds to all active regions in the heatmap. The visualization of $\mathcal{U}$ can be seen in Figure \ref{fig:overview}(b).

After finding all active regions $\mathcal{U}$, we 
compute the number of unconnected areas to determine the number of clusters. To this end,
we consider the 2D connectivity of pixels. Concretely, two pixels are considered neighbors if they have horizontal, vertical, or diagonal connectivity and their activations are the same (i.e., either 0 or 1). While computing the number of unconnected areas, we discard an area if it is very small (experimentally set as less than 150 pixels), and we consider the background to be another region. The calculated number of unconnected regions, $K$, is provided as the number of clusters for the K-means clustering algorithm.

\subsubsection{Using K-Means Clustering}
For some activations in heatmaps, it can be difficult to determine whether close active regions belong to the same cluster. In these cases, the neighboring method explained in Section \ref{numCLus} is unable to separate the active regions efficiently. For instance, in Figure \ref{fig:Clustering}, it is not possible to determine which active area belongs to which cluster by only checking their connectivity. To address this problem, we employ K-means clustering.

In order to cluster each pixel $p$, we consider the following features: $f(p) = \{x_p,\ y_p,\ r_p,\ g_p,\ b_p  \}$. In our formulation, $x_p$ and $y_p$  are the normalized horizontal and vertical coordinates of pixel $p$. $r_p$, $g_p$ and $b_p$ represent the normalized intensity values of the red, green and blue channels.

First, we apply a Gaussian filter to the heatmap $\mathcal{H}$ to smooth the image. The Gaussian kernel's width and height are set as 11, and the smoothed image is represented as $\mathcal{H}_g$. 

We define another set $\mathcal{W}$ such that every element in $\mathcal{W}$ corresponds to a pixel $p$ and contains $f(p)$ if $p$ is active or zeros if $p$ is inactive:
%
\begin{equation}
\mathcal{W} = \{ ||r_p > T_m\ \textrm{or}\ g_p > T_m||, \quad \forall p \in \mathcal{H}_g\},
\label{eqn:clus}
\end{equation}
{\noindent}where $||.||$ sets the value as $f(p)$ when the condition is correct, and 0s otherwise. We set threshold $T_m$ as 0.5 because we do not need to consider regions with low activation.

After finding the number of clusters, $K$, and features for each pixel in $\mathcal{W}$, we cluster $\mathcal{W}$ using the K-means algorithm. 
The centroids of the clusters are initialized randomly 
and they are updated by minimizing the within-cluster sum-of-squares. The maximum number of iterations for the algorithm is set to 300.

After obtaining the clusters from the K-means algorithm, we check whether there are unconnected regions within the same cluster. If a cluster has unconnected regions, we separate these regions into different clusters using 2D neighboring connectivity, as described in Section \ref{numCLus}. Also, we discard a cluster if it is too small ($<$ 150 pixels). Therefore, the total number of clusters can be different than the $K$ value.

We represent all of the obtained clusters as $C$ and each cluster in $C$ as $C_i$ -- see Figure \ref{fig:overview}(c) for visualization of $C$. We calculate the activation of each cluster $C_i \in C$ using the channel intensities in $\mathcal{H}$:
\begin{equation}
a_{C_i} \leftarrow  \dfrac{1}{n_{C_i}}\sum_{\forall p \in C_i} (w_r\times r_p +w_g \times g_p ), \quad  \textrm{for}\ C_i \in C,
\label{eqn:activation}
\end{equation}
{\noindent}where $r_p$ and $g_p$ are the normalized red and green channel intensities in $\mathcal{H}$, and $n_{C_i}$ represents the number of pixels in region $C_i$. Further, $w_r$ and $w_g$ are the activation weights for the red and green channels. We experimentally set $w_r$ as $0.7$ and $w_g$ as $0.3$. $w_r$ has a higher weight than $w_g$ because red channels reflect more about the activation in our heatmap.

After finding activation value $a_{C_i}$ for each $C_i$, we sort the clusters in descending order of their activation levels. We represent these sorted clusters as $C_{sorted}$. For each $C_i \in C_{sorted}$, we obtain the smallest bounding boxes covering $C_i$. The obtained bounding boxes are represented as $B_{sorted}$, and we consider $B_{sorted}$ as the candidate bounding boxes most likely to belong to the described object.


The overall procedure is summarized in Algorithm \ref{alg:all_model}.

\begin{algorithm}[t!]
	\caption{Finding the Described Scene Region.
		\label{alg:all_model}}
        \KwIn{
        a scene and a referring expression.
        }
         \KwOut{
         	$B_{sorted}$, the candidate bounding boxes belonging to the described object.
        }
        Generate the heatmap $\mathcal{H}$ using Grad-CAM for the given scene and the referring expression\\
        Set $\mathcal{U}$ to be the all active regions in $\mathcal{H}$ (Eq. \ref{eqn:num})\\
        Let $K$ to be the number of disconnected areas in  $\mathcal{U}$\\
        Obtain $\mathcal{H}_g$ by applying a Gaussian filter to $\mathcal{H}$\\
        Let $\mathcal{W}$ to contain the feature vectors of pixels in $\mathcal{H}_g$ (Eq. \ref{eqn:clus})\\
        Cluster $\mathcal{W}$ using K-means clustering with $K$ number of clusters\\
        Set $C$ to be the clusters obtained from K-means clustering\\
        Calculate the activation $a_{C_i}$ for each cluster $C_i \in C$ (Eq. \ref{eqn:activation})\\
        Obtain $C_{sorted}$ by sorting $C$ in terms of the cluster activations\\
        Set $B_{sorted}$ to be the smallest bounding boxes covering each cluster in $C_{sorted}$ \\
        Provide $B_{sorted}$ as the candidate bounding boxes belonging to the described object\\
\end{algorithm}

\section{Experiments and Results}
\label{MAttNet}

To assess our method's efficacy, we selected a state-of-the-art referring expression comprehension method as a baseline (i.e., MAttNet~\cite{yu2018MAttNet}), gathered varied real-world images, and compared the results of both methods on these images.

For a given scene and referring expression, MAttNet first obtains the candidate objects using an object detection module. Then, the method checks how well the expression fits each of the candidate objects. Finally, the candidate object that best fits the expression is considered the target object. To compare our method with MAttNet, we sort the candidate bounding boxes by how well they fit the expression. Similar to our output, the bounding boxes ordered from the most likely to the least likely are considered MAttNet's candidate bounding boxes belonging to the described object.

\subsection{Data Collection}

First, we gathered a dataset of 25 images containing indoor and outdoor scenes (12 images from SUN~\cite{xiao2010sun}, 8 images from Google Images, 4 images from Dogan et al.~\cite{dougan2019learning}, and 1 image from SUN-RGBD~\cite{song2015sun}). These images are classified as easy (7 images), medium (8 images), and hard (10 images) difficulty levels. 
An image is labeled as easy if there are only a few objects in total, they are commonly known objects (e.g., bottle, book, mouse, etc.), and the number of same-type objects is 2 (i.e., only one distractor per object). If the objects are common, but the number of distractors is at least three per object, the image is classified in the medium category. The images in the hard group contain many objects with distractors and some objects that are not so common (e.g., radish, papaya, and artichoke). Since MAttNet uses Mask R-CNN~\cite{he2017mask} for extracting objects, we determine an object as common if it is part of the list of instance categories of Mask R-CNN (i.e., 90 types of objects), so a fair comparison is ensured. Next, one target object per image is annotated by a person blind to our research questions (female, 29 years old). She was instructed to draw a bounding box around an object she would consider difficult to describe.




Thereafter, we used Amazon Mechanincal Turk (AMT) to collect written expressions describing the target objects in the images.
We asked AMT workers to provide an unambiguous description of the target object such that it could be differentiated from other similar objects in the image and gave them some examples. We asked them to describe the objects to a robot in order to collect descriptions that simulate interactions between a user and a robot (e.g., a user requests an object from a robot). 
For each interaction, each user could describe an object using its various features or refer to an object in relation to other objects. For example, different AMT workers described the object in Figure \ref{radis} as `the brown vegetable on the top right', `the purple vegetable right next to the mushrooms', and `the turnip to the right of the eggplant'. To account for this variability, we gathered 10 expressions describing the same target object in the same image. In total, we obtained a dataset with 25 images, 25 target objects (one per image), and 250 expressions (ten per image).

We gathered such a dataset to evaluate our method's performance in different conditions. The easy and medium difficulty images represent the typical computer vision datasets for referring expression comprehension (e.g., RefCOCO dataset~\cite{yu2016modeling} which contains MSCOCO~\cite{lin2014microsoft} images where MAttNet and NeuralTalk2 were trained). In these scenes, the total object categories are limited (91 novel object categories for COCO images) and detectable by existing object detectors. On the other hand, in our hard category dataset, the object categories go beyond the existing datasets, and this dataset represents the scenes that can be encountered in the wild. Therefore, this three-level difficulty dataset enables us to observe the behavior of the methods in many interactions at different difficulty levels. Further, neither NeuralTalk2 nor MAttNet were trained on our collected scenes and expressions, which helps us to better evaluate the methods' generalization capacities.

\subsection{Evaluation Procedure}

The collected expressions were used to generate candidate bounding boxes using our method and the baseline. The first three candidates from each method are considered to compute a matching score with the target object bounding box. %
To calculate the matching score, $S_i$, we use $1-L_{DIoU}$, where $L_{DIoU}$ (defined by Zheng et al.~\cite{zheng2020distance}) represents the matching loss function between two bounding boxes. Therefore, $S_i$ is:
\begin{equation}
S_i \leftarrow \dfrac{area(b_i \cap b_{target})}{area(b_i \cup b_{target})}  - \dfrac{d^2}{c^2},
\label{eqn:match}
\end{equation}
\noindent where $b_i$ is the candidate bounding box and $b_{target}$ is the box of the target object. $d$ represents the normalized distance between the centers of $b_i$ and $b_{target}$, and c is the normalized diagonal length of the smallest box covering $b_i$ and $b_{target}$. 

In Eq. \ref{eqn:match}, the first term gives a higher score for a higher intersection of the boxes, and the second term penalizes the distance between their center of masses. The matching score $S_i$ can vary in [-1,1] interval. The first of the three candidates that results in $S_i > 0$ is accepted as the candidate box showing the region belonging to the target object. In the case of all three candidates having a score lower than zero, we report it as none of the candidate boxes belonging to the target object. %
The same steps are applied to both methods for the 250 expressions (25 images, 10 expressions per image).
Both methods could find at least three candidate boxes in all cases, except for MAttNet in one instance. That image belongs to the easy category, and it was able to find the target object for the first two candidates without affecting the reported results.

\subsection{Results}

\begin{figure}[t!]
\subfigure[All images ($p=0.23$).]{
    \begin{tikzpicture}
\begin{axis}[
    legend style={at={(0.2,0.6)},anchor=west},
    legend columns=3,
    height=3.7cm,
    width=8cm,
    ybar,
    ybar=0pt,
    enlarge x limits=0.1,
    ymin=0, ymax=250,%
    bar width=0.3cm,
    xtick=data,
    xticklabels={1st cand., 2nd cand., 3rd cand., none},
    ylabel style={align=center},
    ylabel= \# of times,
    xticklabel style={align=center},
    cycle list name = exotic,
    every axis plot/.append style={fill,draw=none,no markers}
]
\addplot[color=black!60] table[x=condition,y=1]{plots/all.dat};
\addplot[color=black!30, shift={(0.5,0)}] table[x=condition,y=2]{plots/all.dat};
\end{axis}
\end{tikzpicture}
    \label{fig:all_results}}
\subfigure[Easy images ($p<0.001$).]{
    \begin{tikzpicture}
\begin{axis}[
    legend style={at={(0.2,0.6)},anchor=west},
    legend columns=3,
    height=3.7cm,
    width=8cm,
    ybar,
    ybar=0pt,
    enlarge x limits=0.1,
    ymin=0, ymax=70,%
    bar width=0.3cm,
    xtick=data,
    xticklabels={1st cand., 2nd cand., 3rd cand., none},
    ylabel style={align=center},
    ylabel= \# of times,
    xticklabel style={align=center},
    cycle list name = exotic,
    every axis plot/.append style={fill,draw=none,no markers}
]
\addplot[color=black!60] table[x=condition,y=1]{plots/easy.dat};
\addplot[color=black!30, shift={(0.5,0)}] table[x=condition,y=2]{plots/easy.dat};
\end{axis}
\end{tikzpicture}
    \label{fig:easy_results}}
\subfigure[Medium difficulty images ($p=0.11$).]{
    \begin{tikzpicture}
\begin{axis}[
    legend style={at={(0.2,0.6)},anchor=west},
    legend columns=3,
    height=3.7cm,
    width=8cm,
    ybar,
    ybar=0pt,
    enlarge x limits=0.1,
    ymin=0, ymax=80,%
    bar width=0.3cm,
    xtick=data,
    xticklabels={1st cand., 2nd cand., 3rd cand., none},
    ylabel style={align=center},
    ylabel= \# of times,
    xticklabel style={align=center},
    cycle list name = exotic,
    every axis plot/.append style={fill,draw=none,no markers}
]
\addplot[color=black!60] table[x=condition,y=1]{plots/ambiguous.dat};
\addplot[color=black!30, shift={(0.5,0)}] table[x=condition,y=2]{plots/ambiguous.dat};
\end{axis}
\end{tikzpicture}
    \label{fig:ambiguous_results}}
\subfigure[Hard images ($p<0.01$).]{
    \begin{tikzpicture}
\begin{axis}[
    legend style={at={(0.5,-0.35)},anchor=north, /tikz/every even column/.append style={column sep=0.5cm}},
    legend columns=3,
    height=3.7cm,
    width=8cm,
    ybar,
    ybar=0pt,
    enlarge x limits=0.1,
    ymin=0, ymax=100,%
    bar width=0.3cm,
    xtick=data,
    xticklabels={1st cand., 2nd cand., 3rd cand., none},
    ylabel style={align=center},
    ylabel= \# of times,
    xticklabel style={align=center},
    cycle list name = exotic,
    every axis plot/.append style={fill,draw=none,no markers}
]
\addplot[color=black!60] table[x=condition,y=1]{plots/challenging.dat};
\addlegendentry{Our Method}
\addplot[color=black!30, shift={(0.5,0)}] table[x=condition,y=2]{plots/challenging.dat};
\addlegendentry{Baseline}
\draw ([yshift=-1mm]axis cs:1,100) -- (axis cs:1,100) -- node[above, yshift=-1mm]{\small **} (axis cs:4,100) -- ([yshift=-1mm]axis cs:4,100);
\end{axis}
\end{tikzpicture}
    \label{fig:challenging_results}}
\caption{The number of times that the methods generated candidate bounding boxes that matched the target object by difficulty level.}
\label{fig:results}
\end{figure}
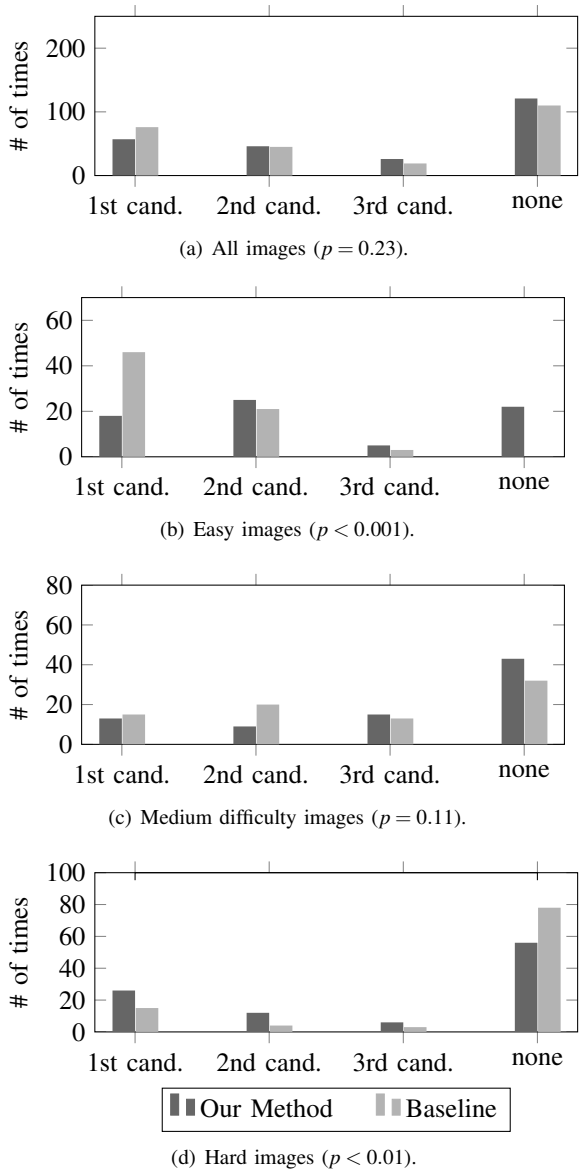

\begin{figure}[t!]
\centerline{
\subfigure[`Man with black pants on the left.']{
	\includegraphics[width=0.20\textwidth,height=0.13\textheight]{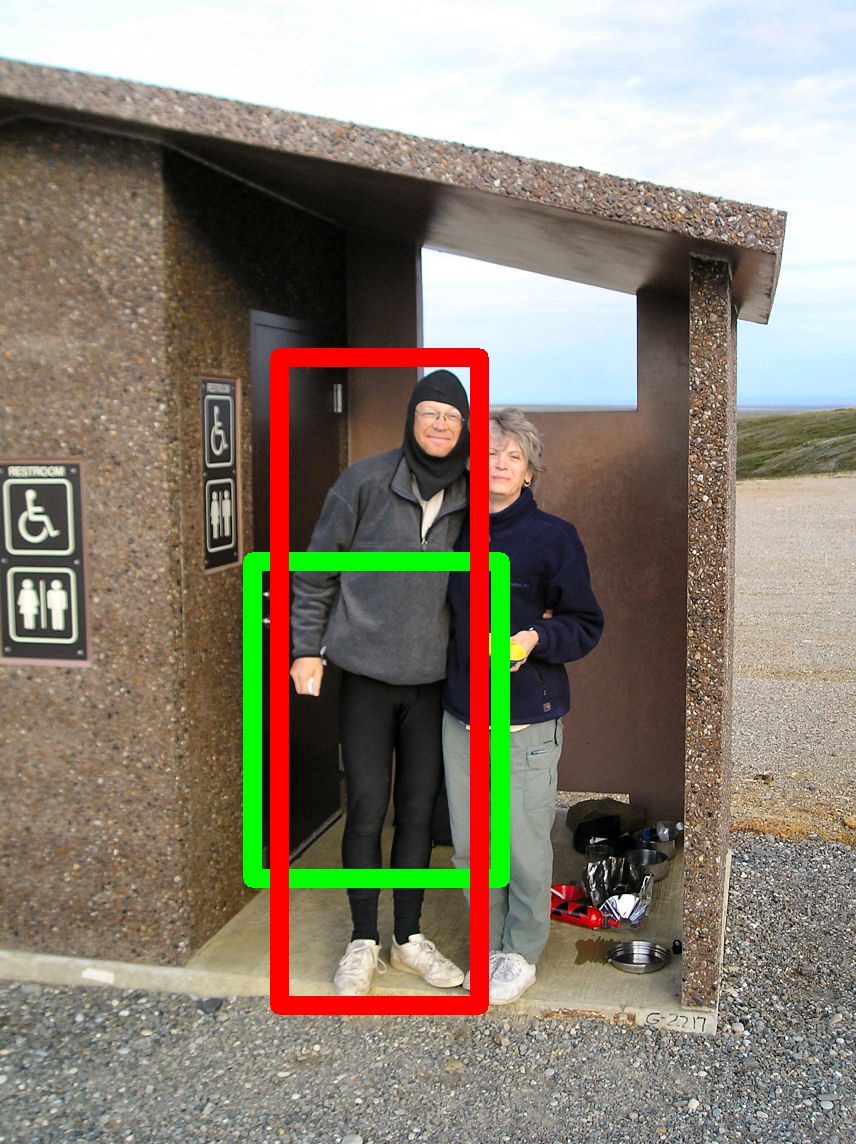}
	\includegraphics[width=0.20\textwidth,height=0.13\textheight]{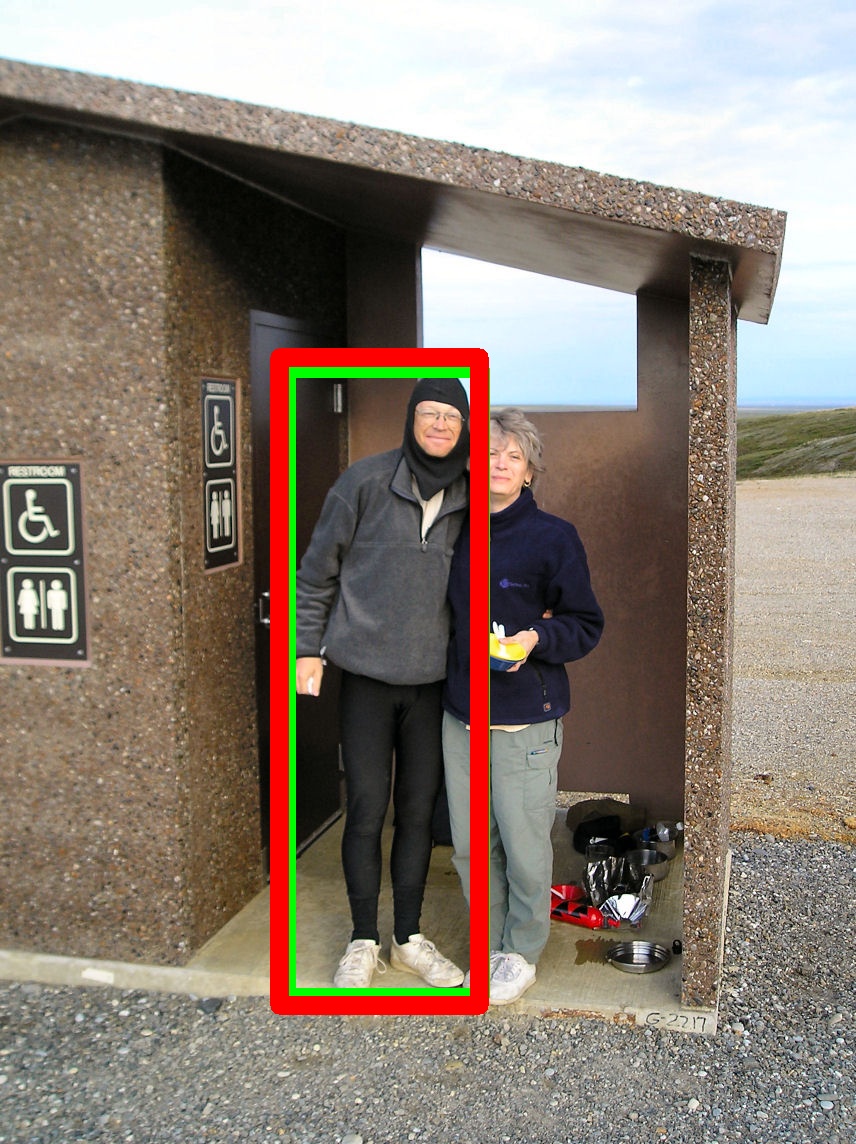}
	\label{man}
	}
}
\centerline{
\subfigure[`The bottle at the back of the table with the yellow top.']{
	\includegraphics[width=0.20\textwidth,height=0.08\textheight]{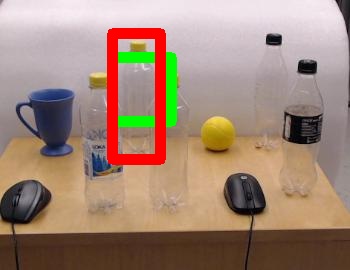}
	\includegraphics[width=0.20\textwidth,height=0.08\textheight]{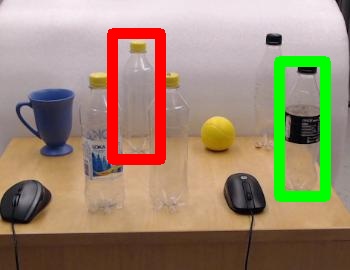}
	}
}
\centerline{
 \subfigure[`The green fruit to the very bottom right in front of the stacks of red apples.']{
	\includegraphics[width=0.21\textwidth,height=0.09\textheight]{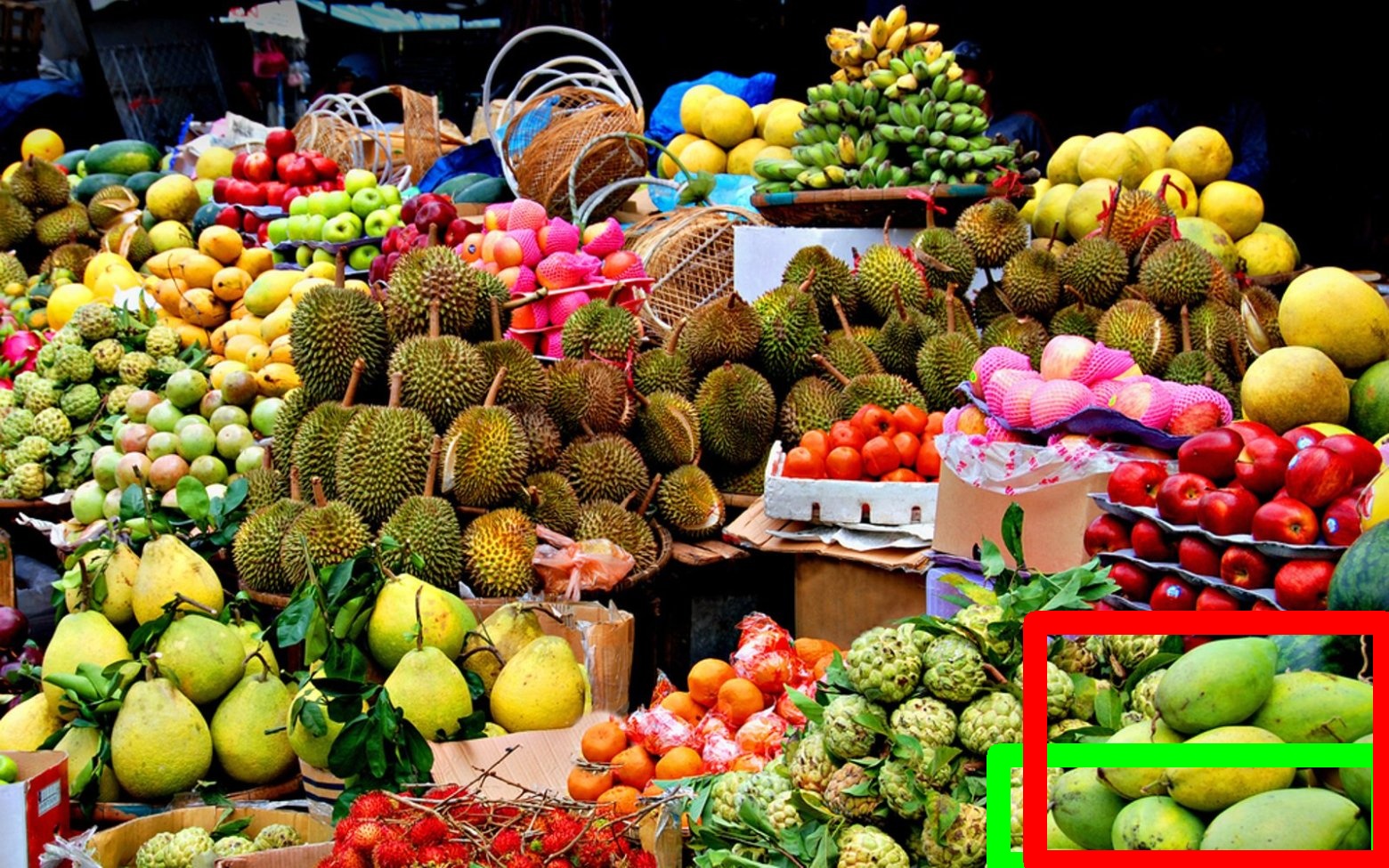}
	\includegraphics[width=0.21\textwidth,height=0.09\textheight]{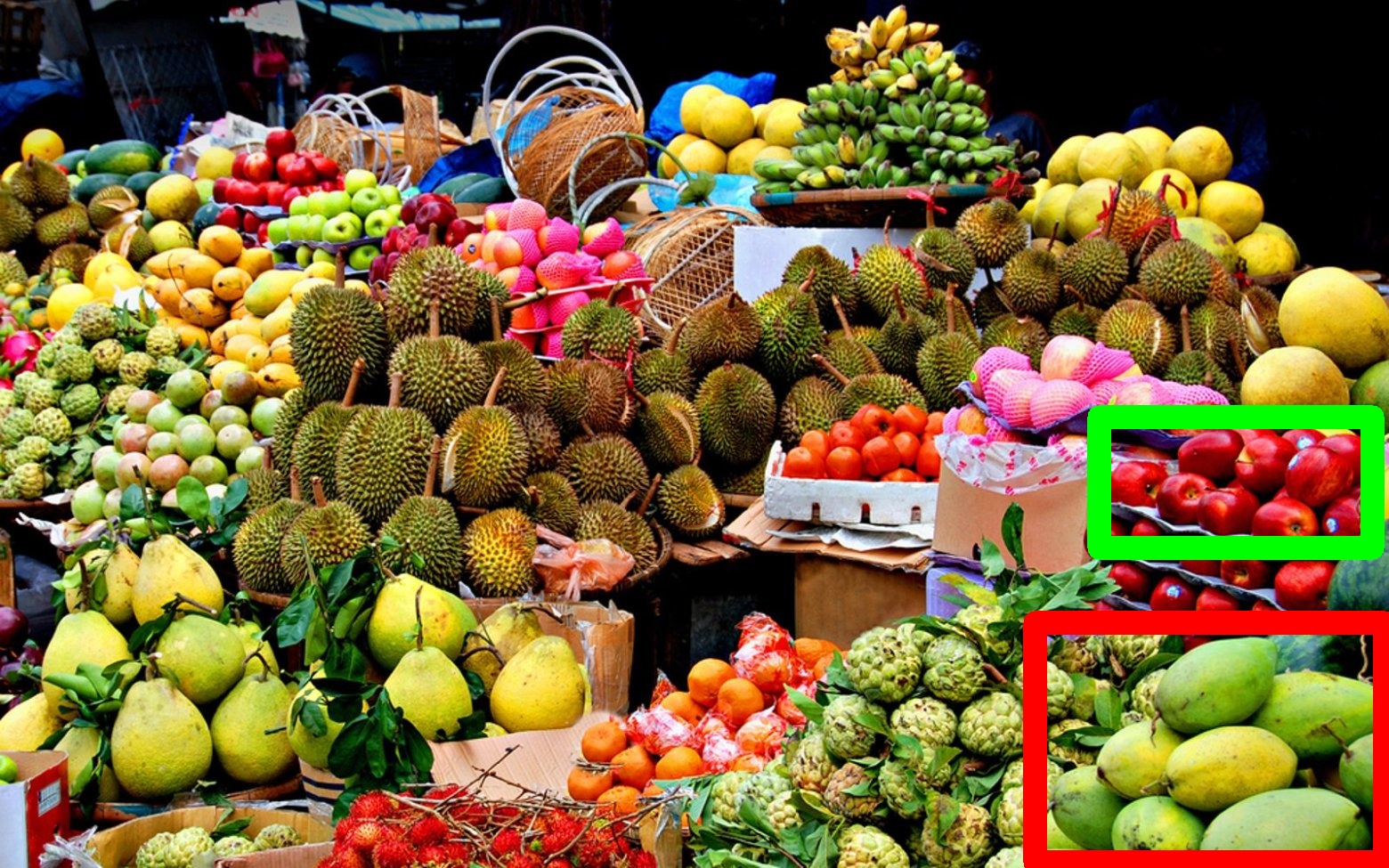}
	\label{papaya}
	}
}
\centerline{
 \subfigure[`The white fish with the yellow stripe just underneath the bigger yellow fish.']{
	\includegraphics[width=0.21\textwidth,height=0.09\textheight]{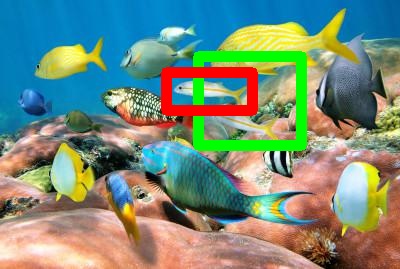}
	\includegraphics[width=0.21\textwidth,height=0.09\textheight]{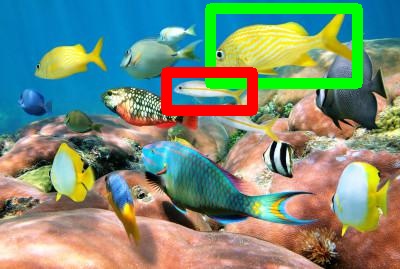}
	\label{fish}
	}
}
\centerline{
 \subfigure[`the brown vegetable on the top right']{
	\includegraphics[width=0.21\textwidth,height=0.09\textheight]{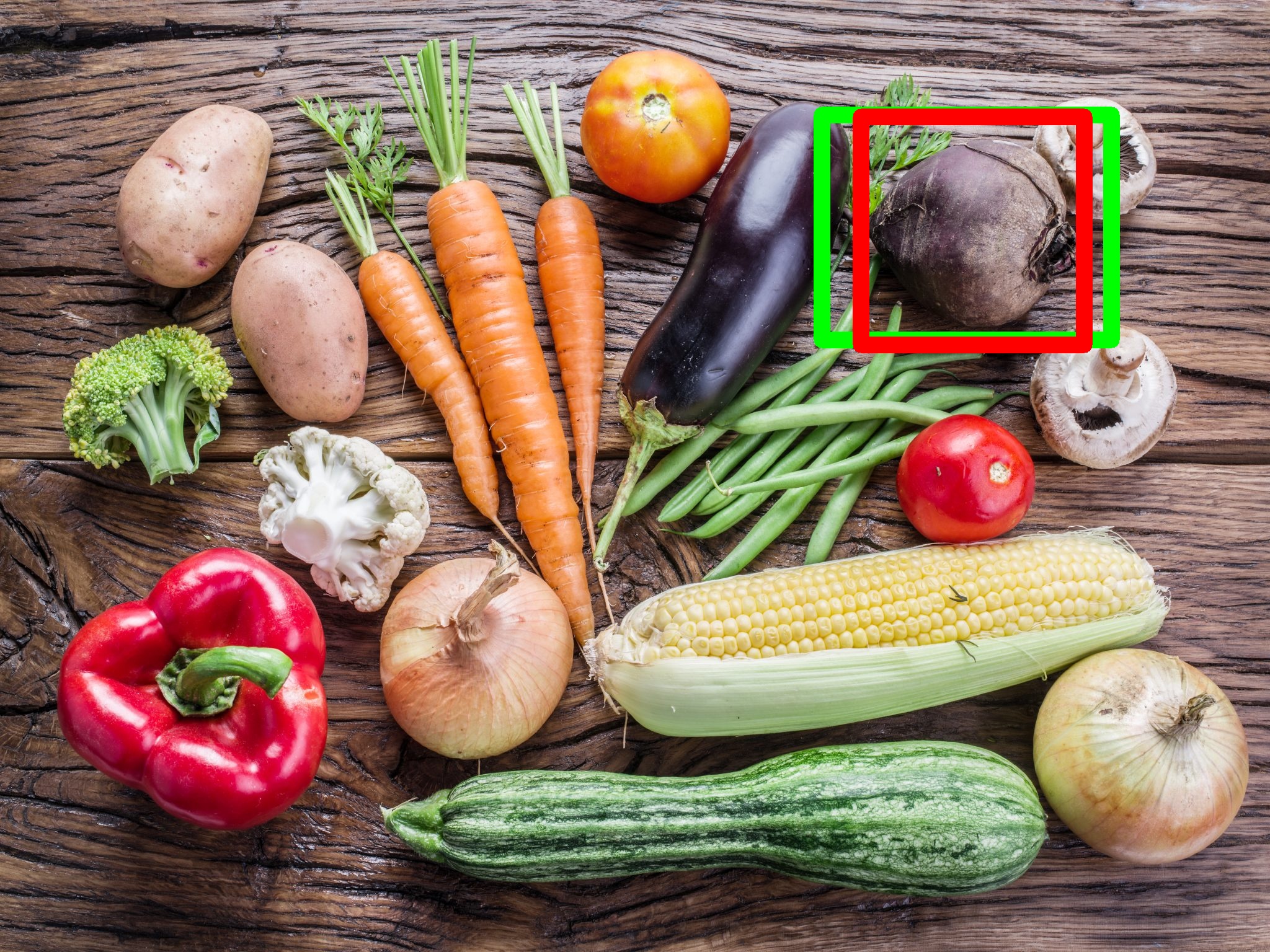}
	\includegraphics[width=0.21\textwidth,height=0.09\textheight]{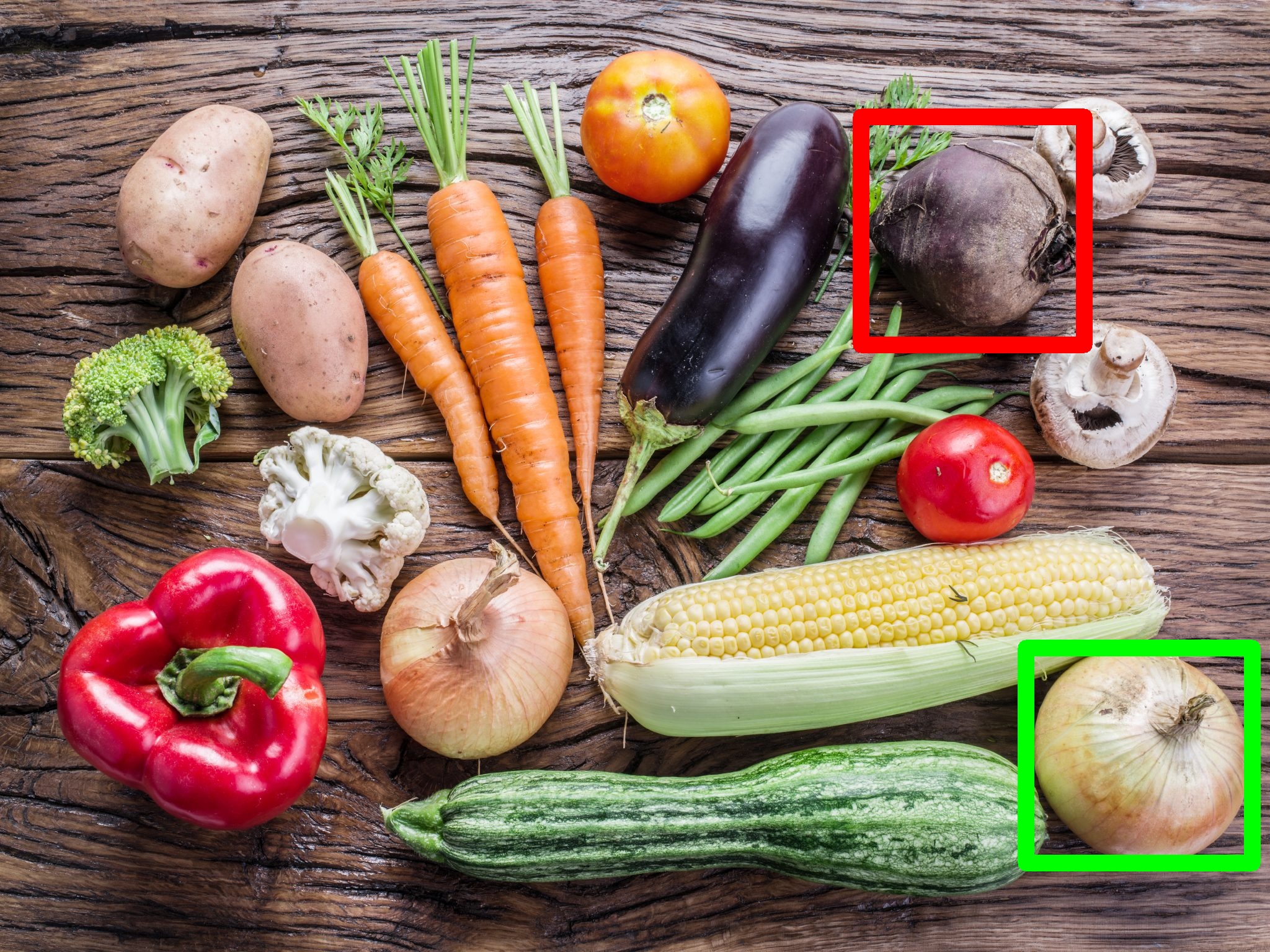}
	\label{radis}
	}
}
\caption{Examples of easy (a), medium (b), and hard (c,d,e) images with original expressions collected from AMT workers describing the target objects in red boxes. The green boxes indicate the first proposed candidate object from our method (left column) and MAttNet (right column). Best viewed in color.}
\label{fig:image_results}
\end{figure}

In this section, we present our results comparing our method with the baseline for 250 expressions (25 images, 10 expressions per image). Figure \ref{fig:results} presents how often the target object matched with the first three candidates for all images at each level of difficulty. 
In Figure \ref{fig:image_results}, we show the first candidates suggested by the two methods for the same images and target objects.



\subsubsection{Results for All Images}

We first compared our method with MAttNet for how many times the target object from the 250 user expressions matched the first, second, or third candidate bounding boxes according to the $S_i$ score from Eq \ref{eqn:match}.
A Chi-Square test did not find any significant differences between the methods, $\chi^2\ (3,\ N = 500) = 4.34,\ p = .23 $. 
Most often, the target object was not matched with any of the first three candidate bounding boxes proposed by the two models (i.e., the mode was ``none'' of the candidates for both methods). In Figure \ref{fig:all_results}, we can see that both methods showed similar trends for different candidates, and the number of times that the methods generated candidate bounding boxes that matched the target object were similar.

\subsubsection{Results by Image Difficulty Level}

First, we examined the results for the easy images with 70 expressions (Figure \ref{fig:easy_results}). We conducted a two-sided Fisher's exact test (the minimum expected value was less than 5 for some cells, so the Chi-Square test couldn't be applied). The results showed significant differences (Fisher's exact test value: $40.29$, $N = 140$, $p < 0.001$. Most often, the target object was matched with the 1st candidate bounding box for MAttNet, and 2nd candidate for our method -- see Figure \ref{fig:easy_results}).
Examining the first candidate, the baseline found the target objects more often than our method did. 
Moreover, there were no cases where none of the baseline's first three candidates was correct, while our method had 22 cases.

For the medium difficulty images, we evaluated the results for 80 expressions. A Chi-Square test did not identify a significant difference between the methods ($\chi^2\ (3,\ N = 160) = 6.07,\ p = .11 $, the mode was ``none'' of the candidates for both methods).
Figure \ref{fig:ambiguous_results} shows that the number of times finding the target boxes was similar for the first and third candidates for both methods. The results from both methods were slightly different for the second and the last items, but these differences were not significant.

Finally, we compared our method with the baseline for the hard category scenes for 100 expressions (Figure \ref{fig:challenging_results}). We again conducted a two-sided Fisher's exact test
, that %
showed significant differences (Fisher's exact test value: $11.44$, $N = 200$, $p = .009$, the mode was ``none'' of the candidates for both methods).
The results indicate that our method found the target object in its first, second, and third candidates more often than MAttNet. Also, the baseline had a higher number of cases for which no candidate was correct.


\section{Discussion}

For easy images, MAttNet performs significantly better than our method. This was expected because there are few objects in the images, the number of distractors per object is only one, and the objects are commonly known. Therefore, the chance level for MAttNet to predict the target is very high (i.e., $1/n$ where $n$ is the total number of detected objects). The chance level is lower for our method because we focus on the activation of each pixel, not the detected object boxes. 

The results for hard images show that
our method performs significantly better than MAttNet at suggesting regions belonging to the target object. This shows that our method can be employed when MattNet fails to identify target objects in challenging environments where there are many objects with distractors and also 
uncommon objects. In these environments, the users mostly referred to the uncommon objects using features such as color, shape, general category (e.g., vegetable instead of radish), and their spatial relationships with known objects nearby. On the other hand, in the easy and medium difficulty images, the users described the objects primarily using the objects' exact names because they are familiar. Therefore, the results indicate that our method performs better than MAttNet when the descriptions are based on an object's features instead of its name.

We did not expect to observe significant differences for the whole evaluation dataset 
and medium difficulty images because our goal in this paper is not an overall performance improvement, given that our method does not simplify the problem to select the target object among the suggested candidates. Instead, we aim to suggest a method that can work better \textit{in the wild} (e.g., with uncommon objects and ambiguities). Therefore, the hard dataset is crucial for the evaluation of such a system. Results on this dataset are critical for human-robot collaboration 
because it is impossible to assume that the robot is familiar with 
all of the different ways that users will use when referring to objects in the real world. In these cases, our method successfully suggests regions by using known concepts. For instance, in Figure \ref{radis}, if the robot doesn't know the concept of a vegetable, it can still predict a region by looking for something brown and on the top right. In other words, our method can handle the unknown objects in the expressions by employing explainability of image captioning and looking for which input features (i.e., which pixels of the image in our case) contribute more to the output.  However, handling unknown objects is more difficult for MAttNet because there should be a detected bounding box to consider an object as a candidate. 

From the qualitative results, 
we observe that our method focuses on the regions which are important for the given expression. For instance,
in Figure \ref{man} from the easy images, our method finds a bounding box focused on the pants of the man because the expression includes this information. Also,  relying on important regions of the scene, not only specified by object categories but also object features, enables our method to handle uncommon objects (e.g., papayas in Figure \ref{papaya}). 
By considering active clusters, we can find regions that better fit expressions than MAttNet can, particularly for hard images.
This is crucial because our goal is to endow robots with the ability to direct their attention to the right part of the scene in the wild and ask for an efficient follow-up clarification instead of asking the user to repeat the whole request again. 

In line with our goal, our qualitative results support that if there are ambiguities in the environment, 
our method can be used to ask for further clarifications by only focusing on the active clusters instead of the whole image. For example, when we asked AMT workers to describe objects as if they are describing them to a robot (i.e., to obtain object descriptions simulating natural language user requests), there were ambiguities in their descriptions. 
For instance, in Figure \ref{fish}, the worker's description fits both of the white fish, and the bounding box obtained from our method contains parts of both fish. In another example, when the description is the green vegetables in Figure \ref{fig:overview}, our method finds the active clusters on the green vegetables for the first two candidates. Also, in Figure \ref{fig:Clustering}, when the red birdhouse is described, our method finds the most active regions on the birdhouses. Therefore, these examples demonstrate that the robot can ask the user to clarify the request by only considering these active regions instead of taking into account the whole images  (e.g. in Figure \ref{fish}, the robot can ask `do you mean the fish on the left or on the right?'). In brief, focusing on active clusters can improve the efficiency of human-robot collaboration.

\section{Conclusion and Future Work}
We propose a method to point the robot's attention in the regions of a scene described by a user to improve human-robot collaboration  in the wild. To achieve this, we find the regions belonging to the described objects using explainability. Our method finds the region activations  using Grad-CAM and then employs K-means clustering to obtain the active clusters. Our qualitative results demonstrate that the regions suggested by our method can be used to resolve ambiguities. Moreover, through our evaluation, we show that our method works better than a baseline  
for scenes with uncommon objects and multiple distractors. 

There could be several extensions of our work. We plan to deploy
our system in a robot and include a human in
the loop to evaluate the efficiency of the interaction while resolving the ambiguities. We also plan to examine this interaction with the perspective of explainable robotics~\cite{setchi2020explainable} considering how users' perception of the robot is affected by the given visual explanations of the system predictions. Further, when our system is deployed in a robot, our method can be expanded by taking into account the aspects of visual attention studies (e.g., the importance of surrounding context~\cite{itti2001computational} or correlation between the visual attention and gaze~\cite{6736067,6180177}). Another promising direction to explore 
could be extending our method to take depth data as a part of its inputs and find the active clusters from 
three dimensions. This could be an important step to use explainability more effectively for human-robot collaboration. 

\section*{Acknowledgment}
This work was partially funded by a grant from the Swedish Research Council (reg. number 2017-05189) and by the Swedish Foundation for Strategic Research. We are grateful to Grace Hung for her voluntary contributions to the data collection, and Liz Carter for her valuable comments.
\bibliographystyle{IEEEtran}
\bibliography{references}


\end{document}